\documentclass[conference]{IEEEtran}
\IEEEoverridecommandlockouts
\usepackage{cite}
\usepackage{amsmath,amssymb,amsfonts}
\usepackage{algorithmic}
\usepackage{graphicx}
\usepackage{textcomp}
\usepackage{hyperref}
\usepackage{xcolor}
\def\BibTeX{{\rm B\kern-.05em{\sc i\kern-.025em b}\kern-.08em
    T\kern-.1667em\lower.7ex\hbox{E}\kern-.125emX}}
\begin{document}

\title{DE-BIASING A FACIAL DETECTION SYSTEM USING VAE\\

}

\author{\IEEEauthorblockN{1\textsuperscript{st} Vedant V. Kandge}
\IEEEauthorblockA{\textit{Dept. of Computer Engineering and Information Technology} \\
\textit{College of Engineering Pune}\\
Pune, India \\
kandgevv18.comp@coep.ac.in}
\and
\IEEEauthorblockN{2\textsuperscript{st} Siddhant V. Kandge}
\IEEEauthorblockA{\textit{Dept. of Computer Engineering and Information Technology} \\
\textit{College of Engineering Pune}\\
Pune, India \\
kandgesv18.comp@coep.ac.in}
\and
\IEEEauthorblockN{3\textsuperscript{st} Kajal Kumbharkar}
\IEEEauthorblockA{\textit{Dept. of Computer Engineering and Information Technology} \\
\textit{College of Engineering Pune}\\
Pune, India \\
kumbharkarkn19.comp@coep.ac.in}
\and
\IEEEauthorblockN{4\textsuperscript{st} Prof. Tanuja Pattanshetti}
\IEEEauthorblockA{\textit{Dept. of Computer Engineering and Information Technology} \\
\textit{College of Engineering Pune}\\
Pune, India \\
trp.comp@coep.ac.in}
}

\maketitle

\begin{abstract}
Bias in AI/ML-based systems is a ubiquitous problem and bias in AI/ML systems may negatively impact society.
There are many reasons behind a system being biased. The bias can be due to the algorithm we are using for our problem or may be due to the dataset we are using, having some features over-represented in it. In the face detection system bias due to the dataset is majorly seen.
Sometimes models learn only features that are over-represented in data and ignore rare features from data which results in being biased toward those over-represented features.
In real life, these biased systems are dangerous to society. The proposed approach uses generative models which are best suited for learning underlying features(latent variables) from the dataset and by using these learned features models try to reduce the threats which are there due to bias in the system. With the help of an algorithm, the bias present in the dataset can be removed. And then we train models on two datasets and compare the results.
\end{abstract}
\section{Introduction}
Often when someone says, the system is biased it refers to a problem where the system is unfairly one-sided on one particular type of input which is over-represented while designing the system. The system with these biases which are used in real-life applications is harmful to society. There are many cases, such as Amazon's attempt to create a resume screening tool that is prejudiced towards women. Amazon’s facial recognition software is also racially prejudiced in 2018. It matches 28 U.S. congresspeople with criminal mugshots. 

In facial recognition systems, there is a problem of racial and gender bias due to the dataset given for face detection. We may be given a dataset with diverse faces and we may not know the correct dissemination of faces in terms of diverse highlights like race, sexual orientation, etc. Due to this, our dataset may conclude up being one-sided on specific occurrences of these highlights that are over-represented within the dataset.

Deep learning-based approaches can be used to solve this problem by mitigating unlabeled spurious correlations by adopting an adversarial debiasing. Models using adversarial debiasing can make much better trade-offs between task accuracy and amplification than other techniques, according to research. Debiasing of the neural network training process is also automated, resulting in faster training convergence and greater accuracy.

The generative models could be used in an effective way to solve this bias problem by learning information irrespective of its type in an unsupervised manner. So in this paper task of facial detection uses generative models to learn latent space which is nothing like the compressed form of data information that can be used to prepare the model for the classification task.

So the paper aims to create a model for facial detection systems and compare the results for two datasets for standard classifiers and generative models for four classes based on race and gender. Furthermore, the related work in this domain is investigated and attempts are made to get better results. The methodology’s goal is to prepare a dataset and perform preprocessing and then create a model based on a generative model. Results are used to compare the outcomes of the two datasets.

\section{Related Work}
Jieyu Zhao, Mark Yatskar, Tianlu Wang, Kai-Wei Chang, Vicente Ordonez.\cite{b1} Presents model leakage and dataset leakage as measures of the intrinsic bias with respect to a protected variable. By using an adversarial debiasing method to remove unwanted features corresponding to protected variables from intermediate representations in the DNN and provide a detailed analysis of its effectiveness. As compared to randomizing model predictions, adversarial debiasing provided a better trade-off and able to mitigate bias amplification by 53 percent to 67 percent while only sacrificing 1.2 to 2.2 points in accuracy.

Brian Hu Zhang, Blake Lemoine, Margaret Mitchell. \cite{b2} Proposed an approach for reducing bias by taking a variable for the group of interest and learning a predictor and an adversary. Text or census data is used as an input to the network X, produces a prediction Y, such as an income bracket or an analogy completion, whereas the adversary tries to model a protected variable Z (such as zip code or gender).

Bernhard Egger, Adam Kortylewski, Andreas Schneider, Andreas Morel-Forster, Thomas Gerig, Thomas Vetter. \cite{b3} Proposed an approach that uses synthetic images of face (highly variable in facial identities and face poses) to analyze and mitigate the damage of dataset bias. It is seen that the damage from dataset bias in real world data can be largely minimized by using synthetic data for pre-training face recognition systems. It depicts that the number of real world face images needed can be minimized by 75\% using synthetic data..

Vikram V. Ramaswamy, Sunnie S. Y. Kim, Olga Russakovsky. \cite{b4} Presents a Generative Adversarial Networks that uses data augmentation method for training fairer attribute classifiers while reducing biases from the correlations between protected attributes (e.g., race, gender) and target labels. This approach highlight that target classifiers trained on the augmented real-world dataset show quantitative and qualitative benefits.

Puspita Majumdar, Richa Singh, Mayank Vatsa. \cite{b5} Proposed an Attention Aware Debiasing methodology that uses an attention mechanism to learn unbiased feature representations by unlearning the model’s dependency on the sensitive attribute. Results of experiment depict that the Attention Aware Debiasing(AAD) approach is able to reduce bias in model prediction and improve the performance of model.

Wilko Schwarting, Guy Rosman, Alexander Amini, Brandon Araki, Sertac Karaman, Daniela Rus. \cite{b6} Introduced a DL based algorithm for end to end autonomous driving that captures variability through an intermediate latent representation. In addition, introduced a novel method for identifying novel inputs which have not been sufficiently trained for by producing the variational autoencoder’s latent uncertainty through the decoder. Ultimately gives an algorithm for debiasing against learned biases based on the unsupervised latent space. By subsampling half of the dataset throughout training and eliminating the over represented latent regions.

Alexander Amini, Ava P. Soleimany, Wilko Schwarting, Sangeeta N. Bhatia, Daniela Rus. \cite{b7} Introduced a debiasing approach to adjust the respective sampling probabilities of individual data points at the time of training. Use this approach to facial detection to promote algorithmic fairness by mitigating hidden biases within training data. Given a biased training dataset, proposed debiased models show increased classification accuracy and decreased categorical bias across race and gender, compared to standard classifiers.

Sixue Gong, Xiaoming Liu, Anil K. Jain. \cite{b8} Presents a de-biasing face recognition network (DebFace) to mitigate demographic bias in face recognition. DebFace learns the disentangled representation for race, age and gender estimation, and face recognition simultaneously. As compared to BaseFace (baseline model), the DebFace representation demonstrates more uniform distribution, which indicates that the demographic information is disentangled from the face representation and faces within different demographic groups are converged together.

\section{Methodology}
\begin{figure}[htbp]
\href{https://www.steveliu.co/vq-vae}{ \centerline{\includegraphics[width=10cm,height=6cm]{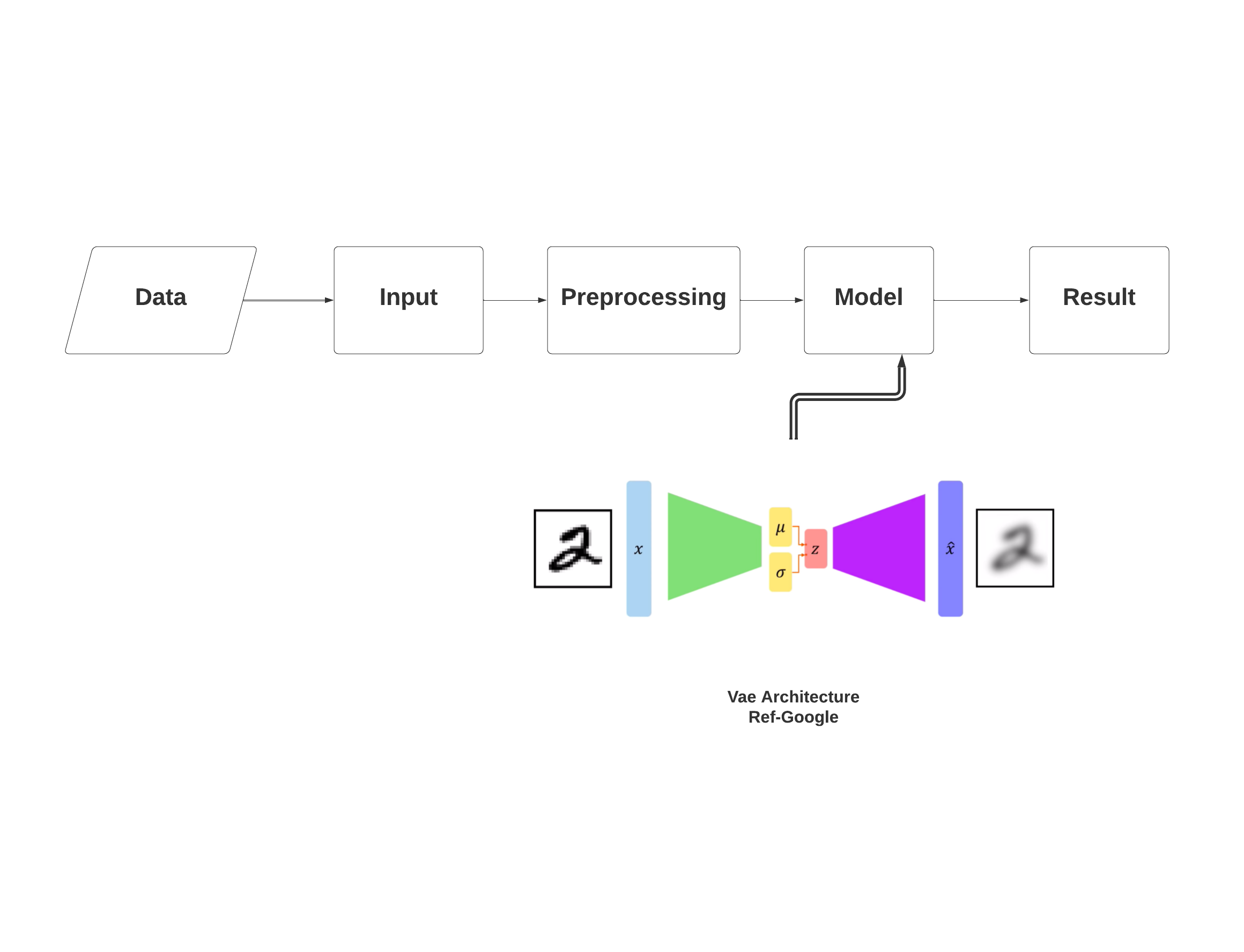}}}
\caption{Flow Diagram}
\end{figure}
Face detection is a binary classification problem with the prediction of the face or not face in the end as result. We have a dataset of images with labels of face or non-face. To achieve results that are not biased on any of the features present in an image such as gender, race or orientation, etc., an underlying latent structure is used and classification is performed. To ensure neutrality in the classification model, datasets created for training should be consistent across all categories of features of the image so that it should be biased on one feature category.

The generative models are the one of best ways to perceive a hidden structure from a dataset in an unsupervised way. By using generative models, latent variables are learned. Latent variables are the compressed form of an image which includes all information of features present in the image. The algorithm uses these latent variables to adaptively resample data while training a model to achieve unbiased results. 

The algorithm uses variational autoencoders in a modified way(called DB-VAE) where it learns latent variables in a completely unsupervised manner and uses it to adaptively resample dataset during training and tries to give unbiased classification. Variational autoencoder is encoder-decoder architecture. An encoder estimate probability distribution q(z|x) and the decoder will do reverse inference to calculate a new probability distribution p(x|z) where z is a latent variable (compressed vector form of an image) that is randomly sampled from a distribution outputted by an encoder network. I.e VAE does not directly learn the latent variable instead it learns a mean and standard distribution of distribution we get a latent variable from it after sampling. It is then sent as input to the decoder which upscales this compressed vector and tries to reconstruct an image similar to the input image. 

As mentioned, latent space in VAE is sampled from the encoder's output and it is not deterministic in nature. So to train the VAE, it uses reparameterization trick where we sample $ \epsilon \rightarrow N(0, 1)$ and derive latent variable (z) as $z = \mu + \sigma . \epsilon $ and this gives z as deterministic form which enable training. 

To train the DB-VAE model architecture it has three loss functions including latent loss, reconstruction loss and classification loss. Latent loss given by KL divergence where it makes sure that difference between learned distribution and prior is minimum. Reconstruction loss which is denoted by L(x,x\^) is calculated as mean squared error and finally the classification loss is standard cross-entropy loss for binary classification.

\begin{figure}[htbp]
\centerline{\includegraphics[width=9cm,height=6cm]{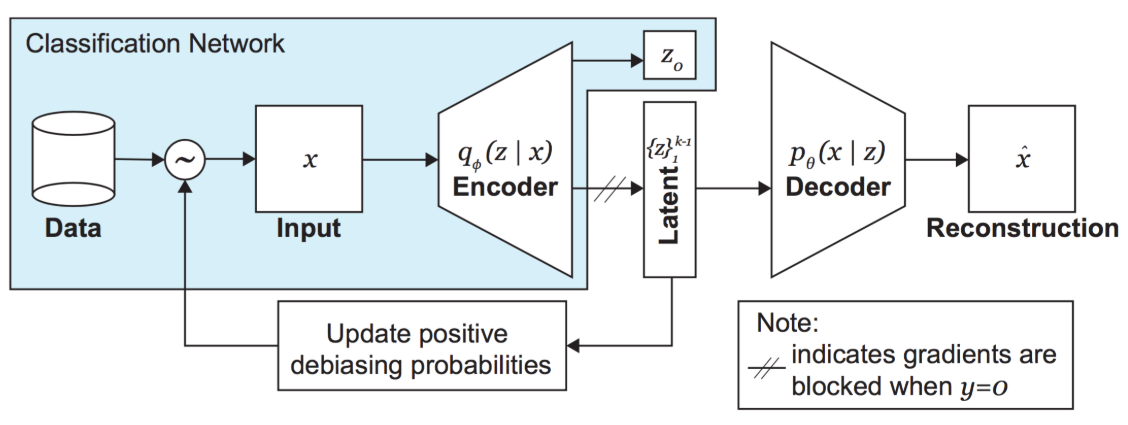}}
\caption{VAE Architecture \cite{b7}}
\end{figure}
Algorithm for debiasing:
The algorithm for debiasing proposed in Amini, Alexander et al.(2019) \cite{b7}is explained in this section. In a dataset for facial detection which have faces and non-face images spread across a wide category such as gender, race, orientation. Some of these features may be over-represented in the dataset which results in the model being biased over these features. 
The algorithm to reduce this bias in system is as follow:
\begin{itemize}
\item VAE network is used to understand hidden underlying features from datasets in an unsupervised manner.
\item The probability distribution for every latent variable is estimated from this learned latent distribution. In this, it may happen that some distributions of these variables are over-represented and some have a lower probability. The probability of selecting an image from a high distribution is likely to be higher than from a lower probability distribution which introduces bias in the system.
\item Then use this inferred distribution to adaptively resample data during training. That is, We'll remodel the likelihood of a specific image being used during training based on how frequently its latent characteristics exist in the dataset. As a result, faces with uncommon traits (such as dark skin, sunglasses, or hats) should be sampled more frequently during training, whereas faces with features that are over-represented in the training dataset should be sampled less frequently (relative to uniform sampling across the training data).
\item This is used to generate a balanced dataset which helps in mitigating bias.
\end{itemize}

\section{Experiments}
A facial detector model is trained on two potentially biased datasets (dataset1 and dataset2 ) using a debiasing algorithm and results are then compared on those two datasets.

Facial detection models while training takes input from a pair including an image and its label. In both datasets,  we have positive and negative examples with positive images having some potential bias for attributes from image such as gender, tone, etc. we train DE-VAE architecture to learn hidden latent structure in positive images and use adaptive resampling as mentioned in the algorithm to debias the model. For negative examples from the dataset, we only train the encoder network. We evaluate the performance of debiased models with standard classifiers with training results of both datasets (dataset1 and dataset2 ) and the PPB dataset.

\subsection{Dataset and Preprocessing}\label{AA}
 
\begin{figure}[htbp]
\centerline{\includegraphics[width=8cm,height=4cm]{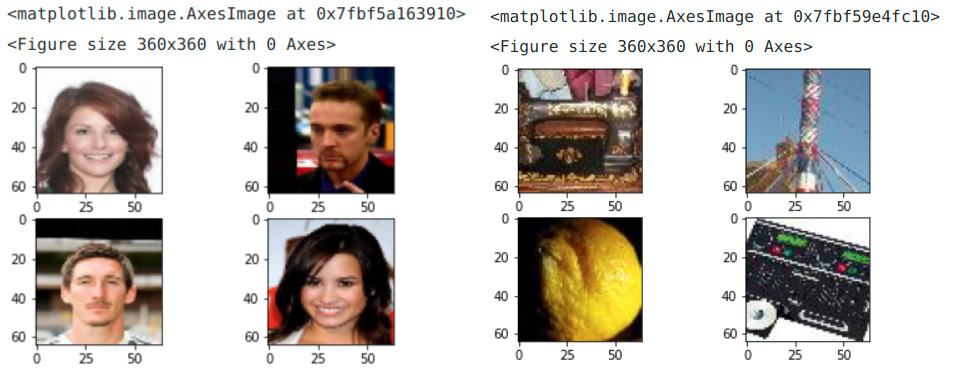}}
\caption{Dataset Examples}
\label{fig3}
\end{figure}
We have used two datasets (dataset1 and dataset2) for our model comparison task. Dataset1 is taken from MIT coursework which contains around 100k images of faces and non-faces. It is already a preprocessed dataset. Dataset2 is a dataset we made from scratch. It has around 300k images. We have used the celebA dataset for face images and for non-face images we have taken a dataset from kaggle and combined it with celebA. For this dataset, we then crop the images to faces and non-face objects and all images are resized to 64x64 size. 

For testing purposes, the PPB dataset is used. This includes photos of 1270 male and female MPs from African and European nations. The dataset has fairness in skin tone and gender, and the images are similar in the pose, lighting, and facial expression. The sex-based "Male" and "Female" labels are used to indicate the gender of each face.

\subsection{Model Architechture and Hyperparameters}

\begin{itemize}
\item To train a facial detection system, we used two different parameter sets and trained models on two datasets.
\item For dataset1, we have 5 layers of sequential convolution with 3 layers with 4x4 kernel size and 2 layers with 3x3 kernel size. In the end, it has 3 fully connected layers with 512, 256 and 1 hidden neuron respectively. 
\item For dataset2, we have 4 layers of sequential convolution with 2 layers with 4x4 kernel size and 2 layers with 3x3 kernel size.  In the end, it has 2 fully connected layers with 1000 and 1 hidden neuron respectively.
\item Then there is a decoder network which mirrors the encoder and tries to reconstruct the input image. Then we train these models to minimize the overall loss function defined above in paper.
\end{itemize}

\section{Results}
On two datasets, we test the performance of our models and test accuracy for both models using the PPB dataset which contains images that are divided into four classes depending on race and gender. So the First dataset contains 110k images of both faces and non-faces taken from the MIT library. While the second dataset is made from scratch and contains around 300k images with both faces and non-faces in a ratio of around 2:1. 
CV2, matplotlib, and TensorFlow libraries are used to perform different preprocessing techniques such as converting images to NumPy array, crop images to the bounding box, and then resizing the images to the required size, etc. on the dataset.
\begin{figure}[htbp]
\centerline{\includegraphics[width=9cm,height=6cm]{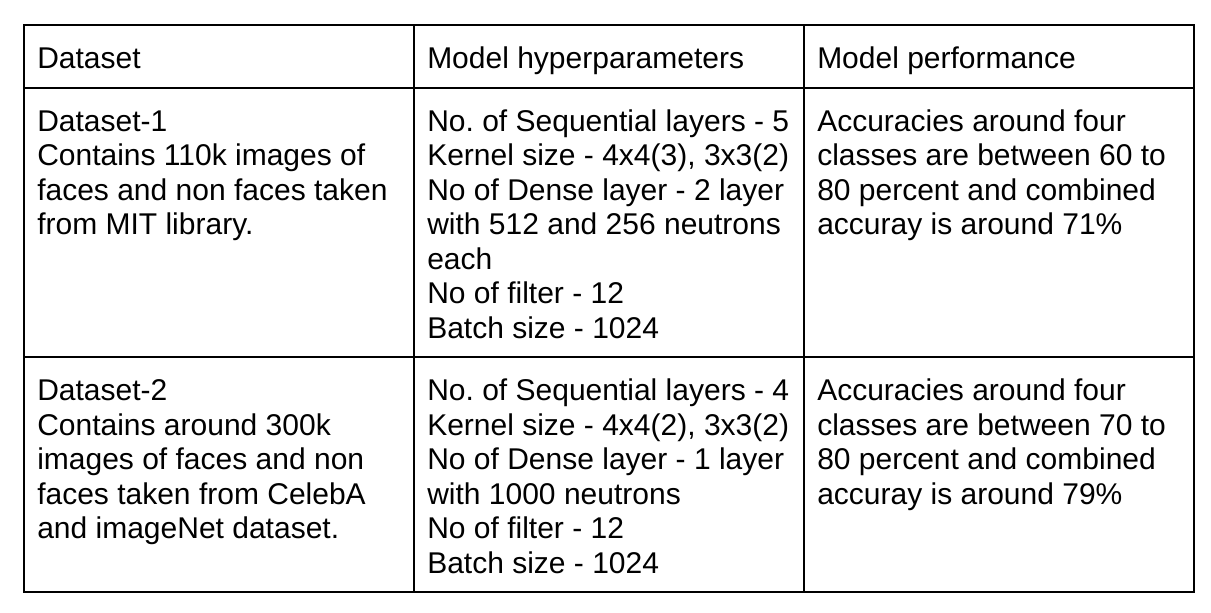}}
\caption{Comparision Table}
\label{fig4}
\end{figure}

Here we observed that hidden latent distribution which contains different facial features such as race, gender, etc can be learned by our models and then passed these encodings to the decoder. So this will prove the hypothesis of the DB-VAE algorithm proposed in Amini, Alexander et al.(2019)\cite{b7} that the algorithm will recognize and then resample the data points which are rarely represented and have a diverse representation of features depending on learned latent features.
To evaluate the performance of the models that we created normalized datasets are used to train the model, for this we created several model architectures and perform hyperparameter tuning, and the after performing training on each architecture we check the performance of each architecture and plotted graph for best of them on test data i.e PPB dataset.
So we trained our model on two datasets which include both faces and non-face images that are given to the model and after training and testing comparison graphs for four classes between standard CNN and VAE-based architectures are plotted.

\begin{figure}[htbp]
\frame{\centerline{\includegraphics[width=8cm,height=4cm]{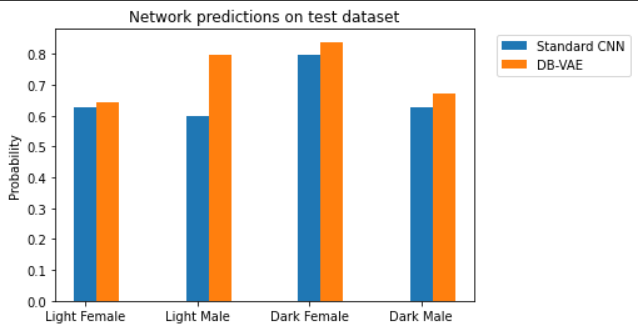}}}
\caption{Results on Dataset 1}
\label{fig5}
\end{figure}

So Fig5 shows the comparison graph for the first dataset as we see there are four bar plots for four classes depending on race and gender (Light female, Light male, Dark female, Dark male) and the plot in blue color shows the performance on standard CNN while the plot in orange color shows the performance on VAE architecture. So as we see VAE architecture performs better along with all four classes for standard CNN but as we can see accuracy for two classes Light male and dark female is slightly greater than the other two classes So here we can say that models perform better than standard CNN but there is still some bias along with the classes.

\begin{figure}[htbp]
\frame{\centerline{\includegraphics[width=8cm,height=4cm]{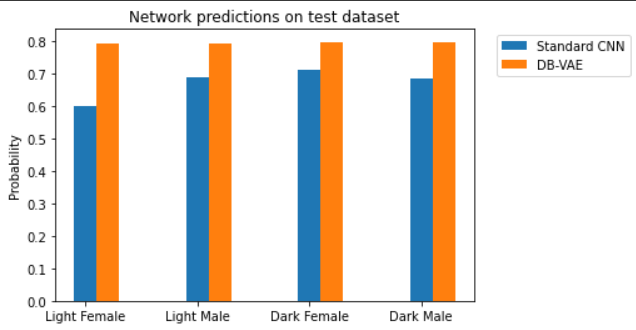}}}
\caption{Results on Dataset 2}
\label{fig6}
\end{figure}
Now, Fig6 shows the comparison graph for the second dataset. As we can see the probability for the VAE architecture for all four classes is greater than the probability for standard CNN and for all four classes probability for the VAE architecture is almost similar. So we can say that the model performs equally along with the four classes i.e. The second dataset outperforms the first in terms of performance. 

\section{Conclusion}
Machine learning models or algorithms might be sometimes biased on certain parameters. i.e., the model may learn just those characteristics that are over-represented in the data, while ignoring uncommon features, resulting in a bias towards the over-represented features.
In this, we created a model architecture that removes biases from the facial detection system and compares the results on two datasets. As we know that we can remove algorithmic bias and detect face is present or not in a given image using generative models and by learning the underlying latent variables in an entirely unsupervised manner.
In comparison to traditional classifiers, our debiased models(DB-VAE) perform better along with all four categories for the second dataset as compared to that of the first dataset.
Here we may conclude that by increasing the size of the dataset and bypassing the biased training data and by using inferred distribution adaptive re-sampling of the dataset during training that images (faces) with rare features should more likely be sampled rather than images with over-represented features we can get more accurate and unbiased results.

\end{document}